\def\year{2021}\relax
\documentclass[letterpaper]{article} 
\usepackage{aaai21}  
\usepackage{times}  
\usepackage{helvet} 
\usepackage{courier}  
\usepackage[hyphens]{url}  
\usepackage{graphicx} 
\usepackage{natbib}  
\usepackage{caption} 
\nocopyright
\title{Crowdsourcing Diverse Paraphrases for Training Task-oriented Bots}
\author {
    Jorge Ram\'{i}rez,\textsuperscript{\rm 1}
    Auday Berro,\textsuperscript{\rm 1}
    Marcos Baez, \textsuperscript{\rm 1}
    Boualem Benatallah, \textsuperscript{\rm 2,1}
    Fabio Casati \textsuperscript{\rm 3} \\
}
\affiliations {
    \textsuperscript{\rm 1} LIRIS -- University of Claude Bernard Lyon 1;
    \textsuperscript{\rm 2} University of New South Wales;
    \textsuperscript{\rm 3} ServiceNow \\
    jorge-daniel.ramirez-medina@univ-lyon1.fr
}

\usepackage{pifont}

\begin{document}

\maketitle

\begin{abstract}
A prominent approach to build datasets for training task-oriented bots 
is crowd-based paraphrasing.
Current approaches, however, assume the crowd  would naturally provide diverse paraphrases or focus only on \textit{lexical} diversity. In this WiP we addressed an overlooked aspect of diversity, introducing an approach for guiding the crowdsourcing process towards paraphrases that are \textit{syntactically} diverse.
%
\end{abstract}

\vspace{-15pt}
\section{Background \& Motivation}

Task-oriented chatbots (or simply bots) enable users to interact with software-enabled services in natural language.
Such interactions require bots to process utterances (i.e., user input) like \textit{``find restaurants in Milan''} to identify the user's intent. 
%
%
%
%
%
A prominent approach to build datasets for intent recognition models involves acquiring an initial set of seed utterances (for the intents) and then grow it by \textit{paraphrasing} this set via crowdsourcing \cite{DBLP:journals/internet/Yaghoub-Zadeh-Fard20}. 

An important dimension to measure quality in this context is \textit{diversity}, i.e., the breath and variety of paraphrases in the resulting corpus, which dictates the ability to capture the many ways users may express an intent.
In this context, paraphrasing techniques generally rely on approaches that aim at introducing \textit{lexical} and \textit{syntactic} variations~\cite{thompson2020paraphrase}. Lexical variations refer to changes that affect individual words, such as substituting words by their synonyms (e.g., \textit{``\underline{search} restaurants in Milan''}). 
Syntactic variations, instead, refer to changes in sentence or phrasal structure, such as transforming the grammatical structure of a sentence (e.g., \textit{``Where can we eat in Milan?''}). 
While the development of techniques to introduce such lexical and syntactic variations is the focus of ongoing work in automatic paraphrasing~\cite{DBLP:journals/pvldb/BerroFBBB21}, they are currently greatly under-explored in the crowdsourcing community.


Among the few contributions towards diversity, a prominent data collection framework involves turning crowd-based paraphrasing into an iterative and multi-stage pipeline. Here, multiple rounds of paraphrasing are chained together, and the seed utterances for a round come from a previous round by using different seed selection strategies (e.g., simply choosing all paraphrases from the previous round \cite{DBLP:conf/lrec/NegriMMGB12}, random sampling \cite{DBLP:conf/acl/JiangKL17}, or identifying outliers \cite{DBLP:conf/naacl/LarsonMLKHLHTM19}).
%
%
The focus of these strategies is to ultimately reduce the bias effect of factors like the seed utterances and examples shown to workers \cite{DBLP:conf/slt/WangBKH12}.
Diversity can be further improved by focusing on the actual crowdsourcing task. This task could constraint the crowd from using frequently-used words \cite{DBLP:conf/emnlp/LarsonZMTCGLK20} or suggest words that workers may incorporate in their paraphrases \cite{DBLP:conf/iui/Yaghoub-Zadeh-Fard20}.
While valuable, these contributions assume workers would naturally produce diverse paraphrases or focus primarily on lexical variations.

In this paper we describe our preliminary work towards a multi-stage paraphrasing pipeline that can guide the crowdsourcing process towards producing paraphrases that are syntactically diverse and balanced.

\begin{figure*}[t]
\centering
\includegraphics[width=1\textwidth]{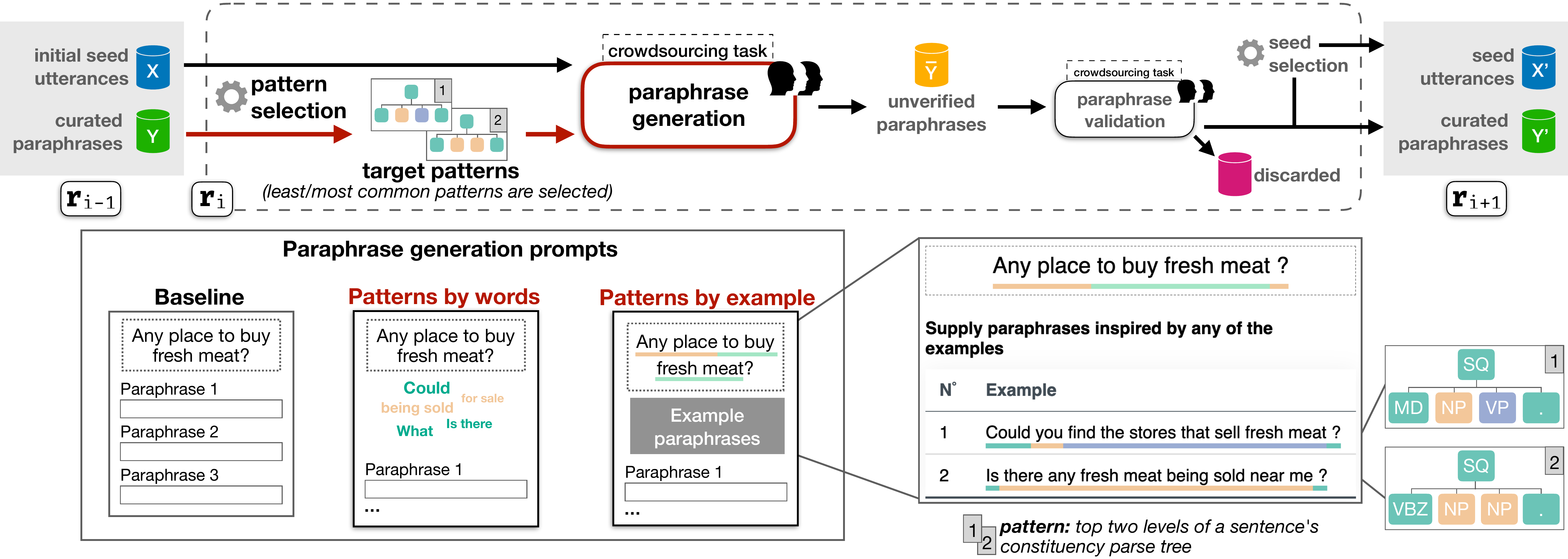}
\caption{
Our approach (in red) sits in a pipeline for paraphrasing. \textit{Pattern selection} identifies target patterns (capturing syntax). 
\textit{Paraphrase generation} leverages these patterns to craft prompts aiming for specific syntax or to elicit novel syntactic variations.
}
\label{fig:approach}
\vspace{-10pt}
\end{figure*}




\vspace{-5pt}
\section{Crowdsourcing Diverse Paraphrases}

Figure \ref{fig:approach} depicts our approach and where it sits in an iterative and multi-stage pipeline for crowd-based paraphrasing based on prior art \cite{DBLP:conf/lrec/NegriMMGB12,kang-etal-2018-data,DBLP:conf/naacl/LarsonMLKHLHTM19}. 
%
%
In this pipeline, a typical round $r$ of data collection (black arrows) takes as input a dataset of seeds utterances $X$ and a curated collection of paraphrases $Y$ (initially, $Y$ can be empty).
The crowdsourcing task in the \textit{paraphrase generation} step asks a worker to provide a set of $n$ paraphrases $y_j$ for an utterance $x$. 
The resulting collection of unverified paraphrases $\bar{Y}$ is fed to the \textit{paraphrase validation} step, where another crowd helps to check for correctness. The correct paraphrases are then appended to the collection of curated paraphrases $Y$.
The \textit{seed selection} step updates (or fully replaces) the seeds in $X$ by sampling from the correct paraphrases to create the set of seeds for the next round.

Our approach assumes an initial ($X$, $Y$) as input and aims to steer the crowd towards specific \textit{patterns} or encourage workers to contribute novel syntactic variations to the input dataset. 
For these goals, we introduce a pattern selection step and propose novel prompts for paraphrase generation.
%

\noindent\textbf{Pattern selection.}
%
%
To capture and control syntax, we follow \cite{DBLP:conf/naacl/IyyerWGZ18} and define a pattern as the top two levels of a constituency parse tree (this depth mostly has clause/phrase level nodes, making syntax comparisons less strict but still effective).
%
%
The pattern selection step thus analyzes the paraphrases in $Y$ and identifies \textit{target patterns} to support the paraphrase generation step towards these goals.

\textit{How to identify target patterns?}
For example, we may choose the $k$ least-frequent patterns in $Y$ as targets, or the $k$ most-frequent ones, any choice informing the generation step differently.
%
Bottom-k patterns may be used to guide workers to provide paraphrases matching any target pattern to collectively contribute more balanced syntax.
While the top-k may be used as ``taboo'' to avoid frequent syntax.



\noindent\textbf{Paraphrase generation.}
Prompts can easily include words and ask workers to avoid/incorporate them in their paraphrases. 
This is not straightforward for patterns, as patterns directly are not informative for non-experts.

\textit{How can the prompts leverage target patterns to steer towards (or encourage novel) syntax?}
To achieve these goals, this work proposes prompts that 1) impose constraints or 2) give recommendations, both by showing example words or paraphrases sampled from target patterns. 
%
%
%
The specific goal shapes the prompt and what the target patterns represent.
For example, if we want to steer the crowd towards uncommon syntax, we can set target patterns to the least frequent patterns in $Y$. The prompts can then show example paraphrases sampled from these patterns and ask workers to contribute paraphrases matching a pattern in any example (i.e., constraining workers to a specific syntax and compensating for less frequent syntax in $Y$).
%
Alternatively, we may aim for novel syntax, so the prompts may use example paraphrases/words as recommendations to inspire workers and encourage them to contribute novel (or ``unseen'') patterns.

\vspace{-5pt}
\section{Ongoing Experiments}




We are running experiments to explore (i) whether our approach can effectively increase syntactic diversity, and (ii) what task designs are more effective for this goal. Below, we overview of our planned experiments.

\noindent \textbf{Datasets}. We selected the ParaQuality dataset \cite{DBLP:conf/naacl/Yaghoub-Zadeh-Fard19}, which contains seed utterances for intents from different domains, including those for Scopus, Spotify, Open Weather, Gmail among other services.


\noindent \textbf{Experimental conditions}\footnote{Screenshots and details at \url{https://tinyurl.com/hcomp2021div}}. We consider six task designs, each representing different prompts. All prompts share the same basic set of instructions.
The \ding{202} \textit{baseline} prompt simply queries for paraphrases for the given seed. Variations of the baseline include \ding{203} \textit{word recommendations} \cite{DBLP:conf/iui/Yaghoub-Zadeh-Fard20} \ding{204} and \textit{taboo words} \cite{DBLP:conf/emnlp/LarsonZMTCGLK20}.
%
%
Our approach  \ding{205} \textit{patterns by example} shows example paraphrases associated with least-frequent patterns and asks workers to use them as inspiration (allowing novel syntax). A variant of this prompt constraints workers to use only patterns present in the examples.
The \ding{206} \textit{taboo patterns} asks for paraphrases with a pattern different than the given example paraphrases (sampled from most-frequent patterns).
We also propose \ding{207} \textit{patterns by words} to show words (sampled from least-frequent patterns) and request workers to use them in their paraphrases.
A variant fixes the position of the words and asks workers to fill in the blanks.
%
Informed by pilots, all conditions include validators to avoid paraphrases that are (clearly) incorrect: (i) check that they are not copies of the examples and are unique after preprocessing (e.g., lemmatizing), and (ii) avoid gibberish, as in \cite{Liu2019OptimizingTD}.



\noindent \textbf{Procedure}. We conduct two full rounds of the pipeline in Figure \ref{fig:approach}, running all conditions. Pattern selection simply counts the frequency of unique patterns using exact matching, and we adopt the approach in \cite{DBLP:conf/naacl/LarsonMLKHLHTM19} for paraphrase validation. Seeds selection is based on random sampling of correct paraphrases.
For the first round ($r_1$), we use the seeds and correct paraphrases from ParaQuality as input.
We recruit English-speaking workers ranked top-20\% in Toloka and collect paraphrases from 10 workers per seed.

\noindent \textbf{Metrics}. We consider commonly-used paraphrase diversity metrics: Type-Token Ratio (TTR), Paraphrase In N-gram changes (PINC) \cite{DBLP:conf/acl/ChenD11}, and DIV \cite{kang-etal-2018-data}.
We also consider a measure of pattern diversity based on \citet{DBLP:conf/acl/JiangKL17}: the number of distinct patterns divided by the total number of paraphrases. 
Following \citet{DBLP:conf/iui/Yaghoub-Zadeh-Fard20}, we also measure the accuracy of an intent detection model trained on the datasets resulting from each condition.

\noindent \textbf{Discussion.} 
We have implemented the pipeline and prompts, informed by pilots, and are ready to start the experiments.

\bibliography{references,refs-local}

\end{document}